\documentclass[]{article}
\usepackage[square,numbers]{natbib}
\usepackage{adjustbox}
\usepackage[toc,page]{appendix}
\usepackage{amsmath}
\usepackage{graphicx} 
\usepackage{subcaption} 
\usepackage{lipsum}
\usepackage{xcolor}
\usepackage{hyperref}

\usepackage{geometry}
\title{\vspace{-1cm} Training for X-Ray Vision: \\ Amodal Segmentation, Amodal Content Completion, and View-Invariant Object Representation from Multi-Camera Video}
\author{
        Moore, Alexander $^*$\\
	\texttt{moore278@llnl.gov}
        \and
        Saini, Amar \footnote{Equal Contribution.}\\
	\texttt{saini5@llnl.gov}
        \and
        Cancilla, Kylie\\
        \texttt{cancilla5@llnl.gov}
        \and
	Poland, Doug\\
	\texttt{poland1@llnl.gov}
        \and
	Carrano, Carmen\\
	\texttt{carrano2@llnl.gov}
}

\begin{document}

\maketitle

\begin{abstract}
Amodal segmentation and amodal content completion require using object priors to estimate occluded masks and features of objects in complex scenes. Recent amodal segmentation work has introduced using temporal features to enrich object representations for improved amodal segmentation and enforce the concept of object permanence and temporal consistency in amodal video segmentation models which modal video object segmentation lacks. Until now, no data has provided an additional dimension for object context: the possibility of multiple cameras sharing a view of a scene. We introduce \textit{MOVi-MC-AC:} \textbf{\textcolor{red}{M}ultiple \textcolor{red}{O}bject \textcolor{red}{Vi}deo with \textcolor{red}{M}ulti-\textcolor{red}{C}ameras and \textcolor{red}{A}modal \textcolor{red}{C}ontent}, the largest amodal segmentation and first amodal content dataset to date. Cluttered scenes of generic household objects are simulated in multi-camera video. MOVi-MC-AC contributes to the growing literature of object detection, tracking, and segmentation by including two new contributions to the deep learning for computer vision world. Multiple Camera (MC) settings where objects can be identified and tracked between various unique camera perspectives are rare in both synthetic and real-world video. We introduce a new complexity to synthetic video by providing consistent object ids for detections and segmentations between both frames and multiple cameras each with unique features and motion patterns on a single scene. Amodal Content (AC) is a reconstructive task in which models predict the appearance of target objects through occlusions. In the amodal segmentation literature, some datasets have been released with amodal detection, tracking, and segmentation labels. However, to date no dataset has provided ground-truth amodal content labels. While other methods rely on slow cut-and-paste schemes to generate amodal content pseudo-labels, they do not account for natural occlusions present in the modal masks. MOVi-MC-AC provides labels for ~5.8 million object instances, setting a new maximum in the amodal dataset literature, along with being the first to provide ground-truth amodal content. The full dataset is available at \href{https://huggingface.co/datasets/Amar-S/MOVi-MC-AC}{https://huggingface.co/datasets/Amar-S/MOVi-MC-AC}
\end{abstract}

\section{Introduction}\label{sec:introduction}
\footnotetext{This work was performed under the auspices of the U.S. Department of Energy by Lawrence Livermore National Laboratory under Contract DE-AC52-07NA27344. Release number: LLNL-JRNL-2007808}

\footnotemark
The ability to conceive of whole objects from glimpses at parts of objects is called gestalt psychology \cite{ozguroglu2024pix2gestaltamodalsegmentationsynthesizing}. The shape and size of object bounding boxes and masks in video may rapidly change as objects undergo changes in position or occlusion through time. Tracking \cite{aharon2022botsort, zhang2022bytetrack}, video object segmentation \cite{cheng2022xmemlongtermvideoobject, cheng2024puttingobjectvideoobject, ravi2024sam2segmentimages}, object retrieval and re-identification \cite{ye2024transformerobjectreid, HAN2025110869}, and video inpainting \cite{lugmayr2022repaintinpaintingusingdenoising, podell2023sdxlimprovinglatentdiffusion} could benefit from consistent object representations which maintain a cohesive object view invariant of occlusion, representation, or perspective change \cite{poland2024seeingobjectsclutteredworld}. Amodal segmentation and content completion are vital in real-world applications of machine learning requiring consistent object understanding and object permanence through complex video such as robots and autonomous driving \cite{cheang2022learning6dofobjectposes, geiger2012, qian2022impdetexploring}. Monocular image amodal segmentation models \cite{fan2023rethinkingamodalvideosegmentation, mohan2022amodalpanopticsegmentation, tran2024aisformeramodalinstancesegmentation, xiao2020amodalsegmentationbasedvisible, ke2021deepocclusionawareinstancesegmentation} rely on object priors to estimate occluded object size and shape through obscurations. Recent monocular video amodal segmentation models \cite{chen2024usingdiffusionpriorsvideo, lu2025taco} use context from temporally distant video features to estimate amodal segmentation masks across time by exploiting object priors of video diffusion models trained on synthetic data. So far, no existing research has investigated using multi-view images and video to generate consistent object representations for the purpose of amodal segmentation. We further develop this research area to introduce multi-view video amodal content completion, a new task in which object visuals are estimated through occlusion using both temporal context as well as multi-view information. We release the first dataset to contain ground-truth amodal segmentation masks for all objects in the scene as well as ground-truth amodal content (or the visible "x-ray" view) of all objects in every scene.

\begin{table}[t]
	\begin{adjustbox}{width=1.3\textwidth,center}
		
		\begin{tabular}{|l|l|l|l|l|l|l|l|l|}
			& \begin{tabular}[c]{@{}l@{}}MOVi-MC-AC\\ (Ours)\end{tabular}    & \begin{tabular}[c]{@{}l@{}}MOVI-Amodal\\ (Amazon)\end{tabular} & SAIL-VOS 3D & SAIL-VOS  & COCOA     & COCOA-cls & D2S    & DYCE      \\
			\multicolumn{9}{l}{\textbf{Statistics}}                                                                                                                                                                                                                                                               \\
			Image or Video                                                                      & Video                                                          & Video                                                          & Video       & Video     & Image     & Image     & Image  & Image     \\
			Synthetic or Real                                                                   & Synthetic                                                      & Synthetic                                                      & Synthetic   & Synthetic & Real      & Real      & Real   & Synthetic \\
			Number of Video Scenes                                                              & 2041                                                           & 838                                                            & 203         & 201       & -         & -         & -      & -         \\
			Number of Scene Images                                                              & 293,904                                                        & 20,112                                                         & 237,611     & 111,654   & 5,073     & 3,499     & 5,600  & 5,500     \\
			Number of Classes                                                                   & 1,033                                                          & 930                                                            & 178         & 162       & -         & 80        & 60     & 79        \\
			Number of Instances                                                                 & 5,899,104                                                      & 295,176                                                        & 3,460,213   & 1,896,296 & 46,314    & 10,562    & 28,720 & 85,975    \\
			Number of Occluded Instances                                                        & 4,089,229                                                      & 247,565                                                        & -           & 1,653,980 & 28,106    & 5,175     & 16,337 & 70,766    \\
			Average Occlusion Rate                                                              & 45.2\%                                                         & 52.0\%                                                         & -           & 56.3\%    & 18.8\%    & 10.7\%    & 15.0\% & 27.7\%    \\
			\multicolumn{9}{|l|}{\textbf{Provided Modalities}}                                                                                                                                                                                                                                                      \\
			Scene-Level RGB Frames                                                              & Yes                                                            & Yes                                                            & Yes         & Yes       & Yes       & Yes       & Yes    & Yes       \\
			Modal Object Masks                                                                  & Yes                                                            & Yes                                                            & Yes         & Yes       & Yes       & Yes       & Yes    & Yes       \\
			Model Object RGB Content                                                            & Yes                                                            & Yes                                                            & Yes         & Yes       & Yes       & Yes       & Yes    & Yes       \\
			Scene-Level (Modal) Depth Masks                                                     & Yes                                                            & Yes                                                            & Yes         & Yes       & No        & No        & No     & No        \\
			Amodal Object Masks                                                                 & Yes                                                            & Yes                                                            & Yes         & Yes       & Yes       & Yes       & Yes    & Yes       \\
			Amodal Object RGB Content                                                           & Yes                                                            & No                                                             & No          & No        & No        & No        & No     & No        \\
			Amodal Object Depth Masks                                                           & Yes                                                            & No                                                             & No          & No        & No        & No        & No     & No        \\
			\begin{tabular}[c]{@{}l@{}}Multiple Cameras\\ (multi-view)\end{tabular}             & Yes                                                            & No                                                             & No          & No        & No        & No        & No     & No        \\
			\begin{tabular}[c]{@{}l@{}}Scene-object descriptors\\ (instance re-id)\end{tabular} & Yes                                                            & Yes                                               & No          & No        & No        & No        & No     & No       
		\end{tabular}
	\end{adjustbox}
	\caption{A comparison of contemporary datasets for amodal video object segmentation, tracking, and amodal content completion. While MOVi-MC-AC does not contain articulating objects, it is by far the largest dataset which provides complete modal masks, amodal masks, and amodal content for all objects in every scene as well as six distinct camera views with unique camera extrinsics and motion patterns with a united object identification label enabling new directions of research in computer vision.}
\end{table}\label{tab:dataset_meta_info}



\begin{figure}[h!]
	\centering
	\begin{subfigure}[b]{0.45\textwidth}
		\centering
		\includegraphics[width=\textwidth]{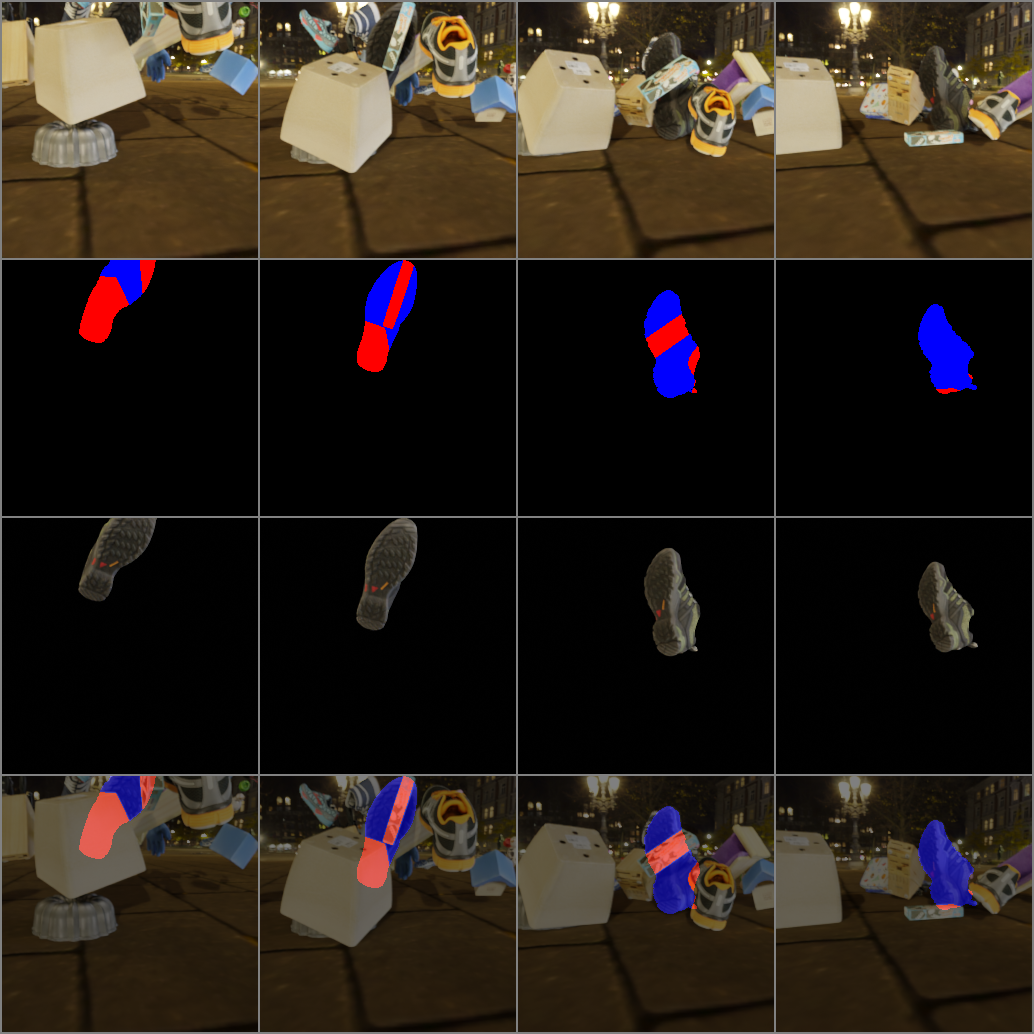}
		\caption{Camera 1.}
		\label{fig:sub_1}
	\end{subfigure}
	\hfill
	\begin{subfigure}[b]{0.45\textwidth}
		\centering
		\includegraphics[width=\textwidth]{images/camera_breakout_1.png} 
		\caption{Camera 6.}
		\label{fig:sub_2}
	\end{subfigure}
	
	\caption{Amodal content completion from multiple cameras must leverage temporal information from one camera view as well as multiple camera perspectives to most accurately predict the visual features of highly-occluded objects from many perspectives simultaneously. MOVi-MC-AC is the first dataset to include ground truth amodal content of occluded objects in video as well as the option to utilize information from multiple cameras on the same scene. Amodal video segmentation and amodal content completion models share information between temporal and camera features to estimate the  unoccluded view of objects. Rows: (1) RGB, (2) Modal/Amodal Masks, (3) Amodal Content (4) Overlay. Each column is a new frame in the video.}
	\label{fig:camera_breakout}
\end{figure}

\begin{figure}[h!]
	\centering
	\begin{subfigure}[b]{0.3\textwidth}
		\centering
		\includegraphics[width=\textwidth]{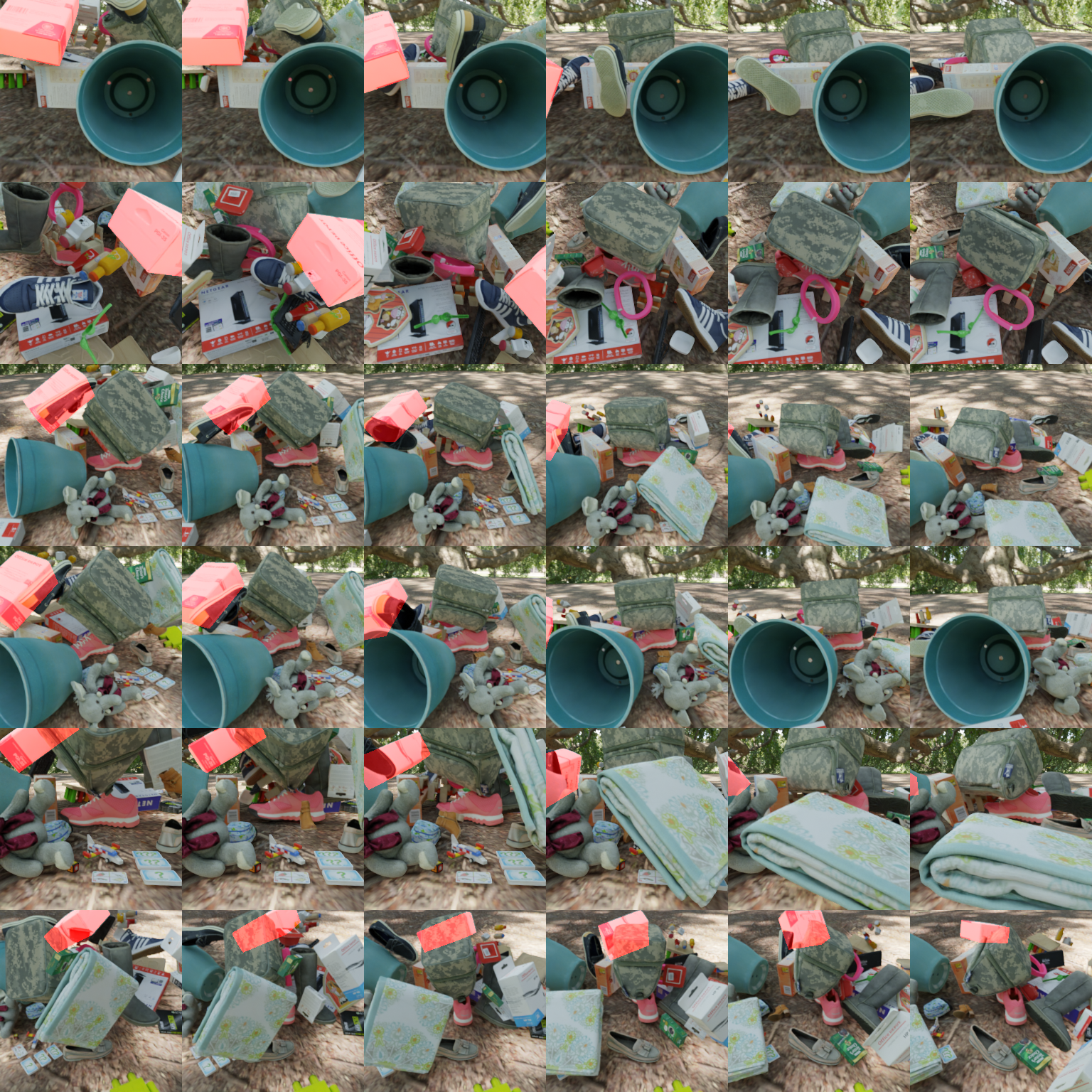}
		\caption{Up to six cameras observe the same scene with different perspectives and motion characteristics.}
		\label{fig:sub1}
	\end{subfigure}
	\hfill
	\begin{subfigure}[b]{0.3\textwidth}
		\centering
		\includegraphics[width=\textwidth]{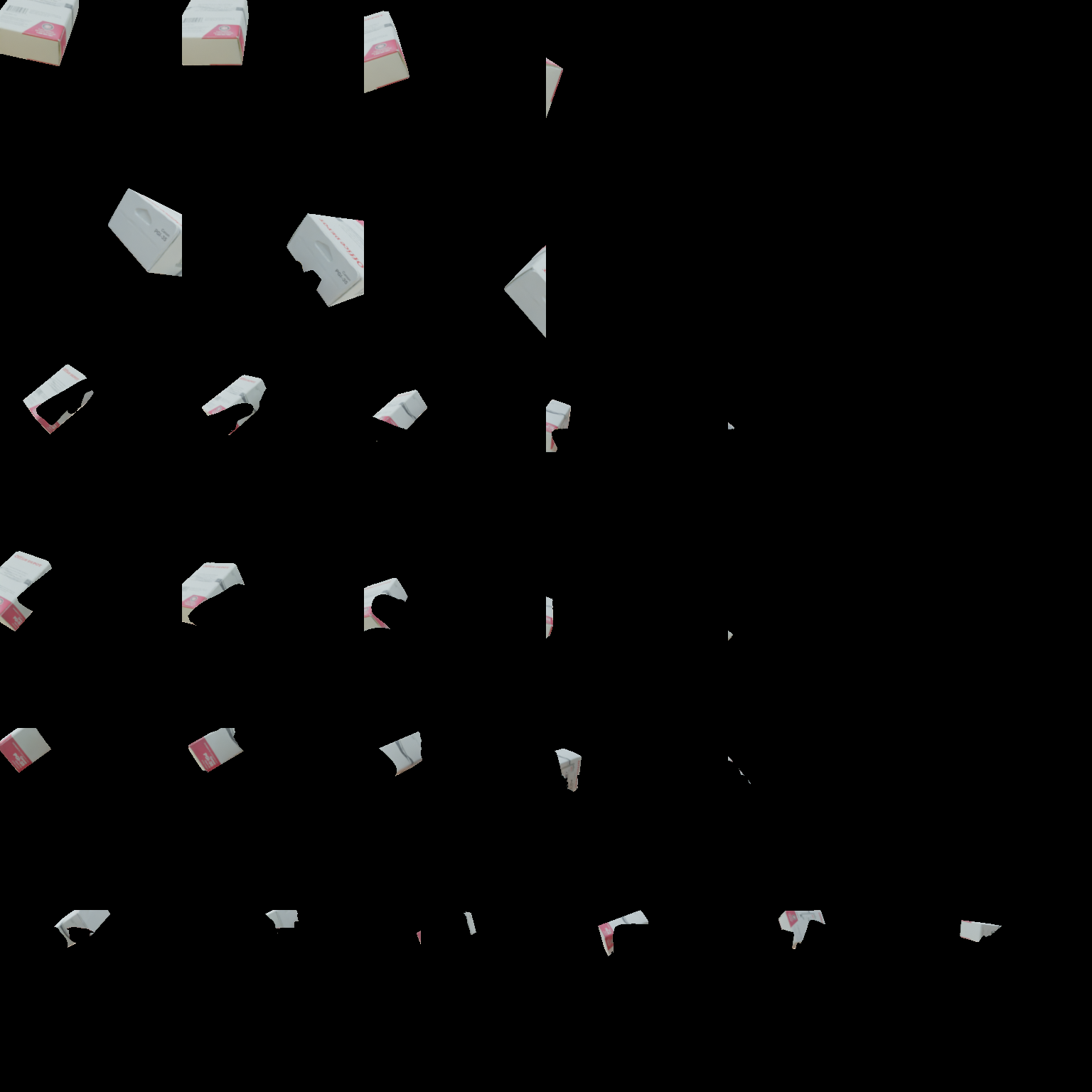} 
		\caption{Object visible features are partially observed by each camera from one perspective.}
		\label{fig:sub2}
	\end{subfigure}
	\hfill
	\begin{subfigure}[b]{0.3\textwidth}
		\centering
		\includegraphics[width=\textwidth]{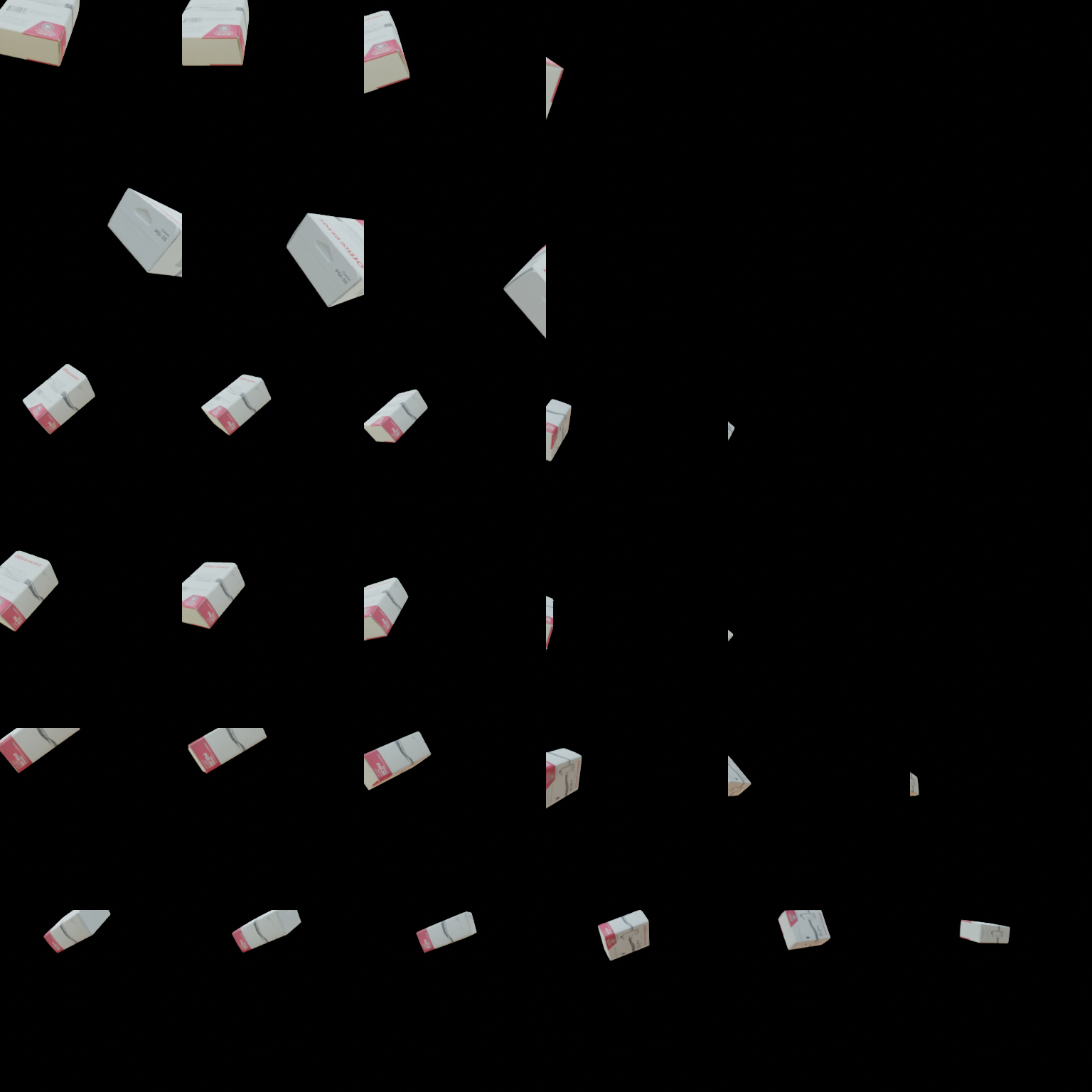} 
		\caption{Amodal object content is the ground-truth unobscured view of the object generated.}
		\label{fig:sub3}
	\end{subfigure}
	
	\caption{Amodal content completion from multiple cameras must leverage temporal information from one camera view as well as multiple camera perspectives to most accurately predict the visual features of highly-occluded objects from many perspectives simultaneously. MOVi-MC-AC is the first dataset to include ground truth amodal content of occluded objects in video as well as the option to utilize information from multiple cameras on the same scene. Amodal video segmentation and amodal content completion models share information between temporal and camera features to estimate the unoccluded view of objects.}
	\label{fig:dataset_splash_demo}
\end{figure}

\subsection{Contributions}\label{sec:contributions}

We make the following contributions to the amodal segmentation and amodal content completion challenge in deep learning for computer vision:

\begin{enumerate}
	\item We release MOVi Multi-Camera Amodal Content, the first dataset to include complete  ground-truth annotations for amodal content, amodal masks, amodal detections, masklets and tracks of obscured objects with over 5 million instances.
	
	\item We introduce the task of multicamera video amodal content completion, including new metrics adapted from the image reconstruction literature to measure the ability of content completion models to correctly predict the shape and appearance of occluded object regions.
\end{enumerate}

\section{Related Work}\label{sec:related}

MOVi-MC-AC is built to fulfill the needs of many tasks in computer vision and object perception for robotics, as well as introduce new tasks to push the field into view-invariant, object-centric representations of objects in cluttered video scenes. We introduce a brief survey of relevant tasks in segmentation and amodal content completion.

\textbf{Image amodal segmentation.} Amodal image segmentation predicts object masks invariant of occlusion \cite{follmann2018learninginvisibleendtoendtrainable, ke2021deepocclusionawareinstancesegmentation, mohan2022amodalpanopticsegmentation, tran2024aisformeramodalinstancesegmentation, tran2024amodalinstancesegmentationdiffusion, xiao2020amodalsegmentationbasedvisible, geiger2012, ke2021deepocclusionawareinstancesegmentation, li2016amodalinstancesegmentation, zhan2020selfsupervisedscene, qi2019amodal_instance_kins}. Image amodal segmentation models require strong object priors to complete object shapes in the absence of temporal context cues. In the presence of temporal context such as real-world video, image amodal segmentation models underperform video amodal segmenters \cite{chen2024usingdiffusionpriorsvideo}.

\textbf{Amodal content completion.} Amodal content completion is relevant to neuroscience in researching how the human mind completes incomplete patterns \cite{doi:10.1177/2041669519840047}. Amodal completion uses object and shape priors in trained diffusion models to estimate occluded content of various kinds \cite{xu2023amodalcompletionprogressivemixed, lugmayr2022repaintinpaintingusingdenoising, podell2023sdxlimprovinglatentdiffusion}. Scene reconstruction \cite{zhan2020selfsupervisedscene} can be performed by decomposing scenes into a collection of de-occluded amodal objects. Extending the object shape priors of image amodal segmentation models, amodal content completion of image and videos requires object appearance priors or temporal context to accurately estimate visual characteristics of occluded objects.

\textbf{Amodal video segmentation.} Contemporary state-of-the-art video object segmentation (VOS) \cite{cheng2022xmemlongtermvideoobject, cheng2024puttingobjectvideoobject, ravi2024sam2segmentimages} segments objects from frame-to-frame through affinity or writing to a memory buffer of references but are prone to identity switching and track loss at low frame rates due to occlusions causing rapid changes in the object mask which modifies the memory encoding \cite{bromley2024addressingissuesworkingmemory}. Amodal video segmentation benefits spatiotemporal stability of object tracking by maintaining consistent object mask representations through video regardless of occlusion \cite{poland2024seeingobjectsclutteredworld, reddy2022, breitenstein2024}.

\textbf{Amodal video content completion.} To the author's knowledge, Diffusion-VAS \cite{chen2024usingdiffusionpriorsvideo} and TACO \cite{lu2025taco} are the only video amodal content completion models currently published. Diffusion-VAS \cite{chen2024usingdiffusionpriorsvideo} uses a three-stage process to estimate depth from monocular video before an amodal segmentation model estimates the amodal mask of the target object using the modal mask and depth before a final pass uses the amodal mask estimate and RGB video to estimate the object amodal content, all powered by a video diffusion model. TACO \cite{lu2025taco} trains a latent video diffusion model to denoise representations of video amodal content in a VAE latent.

\textbf{Multi-camera deep learning.} Multiple-camera paradigms are common in autonomous driving \cite{Sekkat_2022}, robotics \cite{schmied2023r3d3dense3dreconstruction}, and person re-identification \cite{chavdarova2017wildtrackmulticamerapersondataset, han2021mmptrack, AMOSA2023126558, Ferryman2009PETS2009DA} settings. Until now, no dataset has provided full segmentation masks on a multi-camera video corpus. This emphasis on consistent object representations through multiple cameras on the same cluttered scene of generic objects may benefit downstream tasks needing persistence of object representations. Multiple cameras introduces new challenges in object detection, tracking, re-identification as multiple camera perspectives can learn shared representations of objects with view invariance \cite{you2025multiviewequivarianceimproves3d}.

\section{Data}\label{sec:data}

MOVi Multi-Camera Amodal Content (MOVi-MC-AC) is a collection of 2\,041 scenes split into a 1\,651-scene training set and a 390-scene testing set. A scene is a collection of videos containing 6 cameras with unique motion characteristics sampled from static, linear motion, or a moving arc with a camera tracking the middle of the scene \cite{greff2021kubric}. Videos are 2-second simulations in which 24 frames are collected. Each scene contains 2 to 40 objects. Of these, 1 to 20 objects are static on the floor, while 1 to 20 objects are dynamically thrown through the air, sometimes causing extreme obscurations of camera views. The training and testing set contain disjoint sets of objects, enabling re-identification of generic objects to be testable with unseen object classes. Considering all scenes, cameras, and objects, there are approximately 20 million files in this dataset.

\textbf{Annotations.} We provide two kinds of annotations: scene-level and object-level. Scene-level annotations for each camera include the RGB video data collected by the camera, the instance segmentation mask for all objects seen by the camera, and the ground-truth scene depth image. Object-level annotations include the unoccluded amodal RGB content used as the target in amodal content completion, the unoccluded amodal segmentation mask used as the target in amodal image segmentation \cite{ozguroglu2024pix2gestaltamodalsegmentationsynthesizing}, amodal video segmentation \cite{lu2025taco}, amodal instance segmentation \cite{tran2024aisformeramodalinstancesegmentation}, amodal panoptic segmentation \cite{mohan2022amodalpanopticsegmentation}, and the unoccluded depth of the target object which is currently not used as a feature or target in any amodal task. We also provide scene-object descriptors, which associates objects to all relevant scenes that they exist in. Visibility and obscuration rates are also provided within these descriptors.

MOVi-MC-AC is unique among other segmentation datasets where full annotations are provided for \textbf{each} object in each scene. This means MOVi-MC-AC is appropriate for instance and semantic segmentation as well as training model and amodal detectors with information from multiple cameras, whereas VOS and amodal segmentation datasets generally are not (see Section \ref{sec:future_work}) \cite{ravi2024sam2segmentimages, geiger2012}.

Enabled tasks include image segmentation, video object segmentation (VOS), object detection and classification, video object tracking, object re-identification across views or between cameras, and object re-identification across scenes. Along with these common tasks in computer vision research, we also enable new \textbf{amodal} tasks by providing ground-truth labels which no dataset has released until now, including object-based retrieval using amodal content, amodal object detection, amodal 3D detection, amodal video object tracking, multicamera amodal segmentation, and amodal content completion.

Our provided scebe-object descriptors enables development of object re-id \& retrieval pipelines. Finally, from the MOVi dataset engine, we also have access to object names and meta-class categories with descriptions, further supporting research in grounded/referring tracking, and video object segmentation as in language-based detection \cite{cheang2022learning6dofobjectposes}.

Table \ref{tab:dataset_meta_info} compares contemporary datasets relevant to amodal segmentation. We note that no dataset until now supports amodal content prediction natively. Contemporary models use synthetic dataset generation through layering occluded object masks over the target object until a desired occlusion level is achieved \cite{chen2024usingdiffusionpriorsvideo}, \textbf{MOVi-MC-AC is the first dataset to provide ground-truth amodal content}.

\section{Metrics}\label{sec:metrics}

To accompany our new proposed task of multiple-camera video object amodal content prediction, we introduce the following metrics derived from contemporary amodal video object segmentation and computer vision image reconstruction literature.

\subsection{Amodal Segmentation Metrics}\label{sec:amodal_seg_metrics}
Amodal segmentation accuracy can be quantified by segmentation metrics between the predicted and ground truth amodal mask of an occluded object.

Following common practice in amodal segmentation, we propse using the mIoU and mIoU\textsubscript{occ} as evaluation metrics for amodal mask predictions \cite{fan2023rethinkingamodalvideosegmentation, gao2023coarsetofineamodalsegmentationshape, yao2022selfsupervisedamodalvideoobject}. Given videos including the modal and amodal masks for the target object, where the ground-truth modal mask is $M_i$, and the predicted and ground-truth amodal masks are $\hat{A}_i$ and $A_i$, the mIoU metric is given as:

\[
\text{IoU} = \frac{\hat{A}_i \cap A_i}{\hat{A}_i \cup A_i}
\]

and mIoU\textsubscript{occ} as:

\[
\text{mIoU}_{\text{occ}} = \frac{(\hat{A}_i - M_i) \cap (A_i - M_i)}{(\hat{A}_i - M_i) \cup (A_i - M_i)}
\]

following Diffusion-VAS \cite{chen2024usingdiffusionpriorsvideo}.


\subsection{Image Reconstruction Metrics}
MOVi-MC-AC is the first dataset to provide ground-truth amodal content labels for direct training on occluded content reconstruction. In order to measure content completion accuracy of models, we will adapt the following existing metrics from image reconstruction:

\textbf{PSNR metric.} The Peak Signal-to-Noise Ratio is a log-distance between the mean squared error of a target image $x_1$ and its predicted reconstruction $x_2$ for images scaled to a range such that the maximum is $v$:

\begin{equation}
	PSNR(x_1, x_2) = 10 \log_{10} \frac{v^2}{MSE(x_1, x_2)}
\end{equation}

\textbf{LPIPS metric.} The Learned Perceptual Image Patch Similarity \cite{zhang2018unreasonable} uses VGG or AlexNet as embeddings to extract features at multiple scales then measures the distance between the target $x_1$ and reconstruction $x_2$ for activations at layer $l$:

\begin{equation}
	LPIPS(x_1, x_2) = \sum_{l}^{} \lvert f_l(x_1)  - f_l(x_2) \rvert^2
\end{equation}
		
\textbf{SSIM metric.} Structural Similarity Index Measure \cite{wang20024} compares the mean luminance, contrast, and structure between two images, where x and y are the two image patches being compared (e.g., a reference image and a distorted image). SSIM uses \(x\) and \(y\) are the two image patches being compared (e.g., a reference image and a distorted image), \(\mu_x\) and \(\mu_y\) are the mean luminance of \(x\) and \(y\), respectively, \(\sigma_x^2\) and \(\sigma_y^2\) are the variances of \(x\) and \(y\), \(\sigma_{xy}\) is the covariance between \(x\) and \(y\), and \(C_1\) and \(C_2\) are small constants to stabilize the division when the denominator is close to zero:
\begin{equation}
	\text{SSIM}(x, y) = \frac{(2\mu_x\mu_y + C_1)(2\sigma_{xy} + C_2)}{(\mu_x^2 + \mu_y^2 + C_1)(\sigma_x^2 + \sigma_y^2 + C_2)}
\end{equation}

\subsection{Occluded Segmentation and Reconstruction Metrics}
For each image reconstruction metric, we apply the function on  the ground-truth amodal mask and estimated amodal mask, or the ground truth amodal content and the estimated amodal content. In addition to the entire amodal mask and content, amodal segmentation literature also introduces mIOU\textsubscript{occ}, the segmentation performance taken only in the occluded regions to prevent influence of the easy-to-predict visible mask. The suffix ``\textsubscript{occ}" applied to each of our amodal segmentation and amodal content completion metrics means to first subset the  amodal prediction to only the occluded region using the modal mask, before applying the metric on the occluded regions of the ground truth and prediction.

Given amodal mask $A$ and modal mask $V$, the occluded mask $O$ is given by $A\backslash V$. Using $O$ we can generate subimages $I_1'$ and $I_2'$ by applying the ground-truth occluded mask over the ground-truth amodal object content:

\begin{equation}
	\begin{split}
		I_1' &= I_1 \cdot O \\
		I_2' &= I_2 \cdot O
	\end{split}
\end{equation}

Using these two masked images, we can compute ground-truth amodal content quantitative metrics on occluded regions:
\begin{equation}
	\mathrm{Metric_{occ}} = \mathrm{Metric}(I_1', I_2')
\end{equation}
\section{Future Work}\label{sec:future_work}

MOVi-MC-AC enables a wide range of new tasks in computer vision for the detection, tracking, and segmentation of multiple objects across camera views in cluttered scenes. We propose the following tasks as open challenges enabled by MOVi-MC-AC, which could not be directly trained for until now:
\begin{enumerate}
	\item \textbf{Multi-camera object detection and tracking.} Multiple Object Tracking (MOT) algorithms could be adapted to detect an object in multiple camera views simultaneously with the benefit of greater spatial and viewpoint contexts and assign it a consistent unique object id. Then persistently track the object through the multiple views through video.
	
	\item \textbf{Multi-scene object retrieval.} Given an attention prompt (such as a bounding box or segmentation mask) in one video or one multi-camera scene, retrieve the object as a detection in a new camera view of the same scene to unite the camera perspectives. One could further detect the object in a new scene with new cameras and object clutter context. This equates to a view-invariant object representation learnable through multiple cameras to retrieve the object from a gallery of new view points in cluttered scenes.
	
\end{enumerate}

\nocite{*}

\bibliography{main}

\end{document}